%% file: main.tex
\documentclass[conference, letterpaper, twocolumn]{IEEEtran}
\IEEEoverridecommandlockouts
\usepackage[colorlinks=false]{hyperref}
\usepackage{amsmath,amssymb,amsfonts}
\usepackage{algorithmic}
\usepackage{graphicx}
\usepackage{textcomp}
\usepackage{enumitem}
\usepackage[english]{babel}
\usepackage[latin1]{inputenc}
\usepackage[nolist]{acronym}
\usepackage{amsmath}											
\usepackage{amsfonts}											
\usepackage{blindtext} 											
\usepackage{booktabs}											
\usepackage{fancyvrb}
\usepackage{ifthen}												
\usepackage{filecontents} 										
\usepackage[T1]{fontenc} 
\usepackage{graphicx}
\usepackage[dvipsnames,table]{xcolor}
\usepackage{ifthen}												
\usepackage{listings}											
\usepackage{lipsum}
\usepackage{multirow}
\usepackage{pgfplots}
\usepackage{siunitx}	
						
\usepackage{tabularx}
\usepackage{url}
\usepackage{tikz}
\usepackage{circuitikz}
\usepackage{comment}
\usepackage{footnote}
\usepackage{float}
\usepackage{listings}
\usepackage{setspace}
\usepackage{makecell, tabularx}
\usepackage{ragged2e}

\usetikzlibrary{calc,trees,positioning,arrows,chains,shapes.geometric,%
	decorations.pathreplacing,decorations.pathmorphing,shapes,%
	matrix,shapes.symbols}

\usepackage[font=footnotesize,labelfont=bf]{subcaption}		
\usepackage{caption}
\usepackage[ruled,linesnumbered]{algorithm2e}
\SetAlCapNameFnt{\footnotesize}
\SetAlCapFnt{\footnotesize}
\SetAlgoNlRelativeSize{0}
\RestyleAlgo{ruled}
\SetKwComment{Comment}{/* }{ */}

\usepackage[style=ieee, backend=biber, natbib=true, maxbibnames=1, minbibnames=1]{biblatex}
\lstdefinestyle{Python}{
	language        = Python,
	basicstyle      = \ttfamily,
	keywordstyle    = \color{blue},
	keywordstyle    = [2] \color{teal}, 
	stringstyle     = \color{green},
	commentstyle    = \color{red}\ttfamily
}

\usepackage{fancyhdr}

\fancypagestyle{firstpage}{%
	\fancyhf{}%
	\fancyhead[C]{\normalsize To appear at the 20th International Conference on Synthesis, Modeling, Analysis and Simulation Methods, and Applications to Circuit Design (SMACD'24), Jul 2-5 2024, Volos, Greece}
	
	\fancyfoot[C]{\footnotesize \copyright2024 IEEE. Personal use of this material is permitted. Permission from IEEE must be obtained for all other uses, in any current or future media, including reprinting/republishing this material for advertising or promotional purposes, creating new collective works, for resale or redistribution to servers or lists, or reuse of any copyrighted component of this work in other works.}
}

\newboolean{blindreview}
\setboolean{blindreview}{false}
\DeclareRobustCommand{\IEEEauthorrefmark}[1]{\smash{\textsuperscript{\footnotesize #1}}}

\usepackage[a4paper, total={184mm,239mm}]{geometry}

\makeatletter
\def\ps@IEEEtitlepagestyle{%
	\def\@oddfoot{\mycopyrightnotice}%
	\def\@oddhead{\hbox{}\@IEEEheaderstyle\leftmark\hfil\thepage}\relax
	\def\@evenhead{\@IEEEheaderstyle\thepage\hfil\leftmark\hbox{}}\relax
	\def\@evenfoot{}%
}
\def\mycopyrightnotice{%
	\begin{minipage}{\textwidth}
		\centering 
		\footnotesize    
		\copyright2024 IEEE. Personal use of this material is permitted. Permission from IEEE must be obtained for all other uses, in any current or future media, including reprinting/republishing this material for advertising or promotional purposes, creating new collective works, for resale or redistribution to servers or lists, or reuse of any copyrighted component of this work in other works.
	\end{minipage}
}
\makeatother

\begin{document}	

\acrodefplural{ADC}[ADC]{Analog-to-Digital Converters}
\acrodefplural{SoCs}{System-on-Chips}
\acrodefplural{IC}[IC]{Integrated Circuits}
\acrodefplural{IP}[IP]{Intellectual Properties}
\acrodefplural{HDLs}{Hardware Description Languages}
\acrodefplural{CSRs}{Control and Status Registers}
\begin{acronym}[placeholder]
	
	\acro{SoC}{System-on-Chip}
	\acro{CRV}{Constrained Random Verification}
	\acro{CDG}{Coverage Directed Test Generation}
	\acro{ASIC}{Application-Specific Integrated Circuit}
	\acro{IP}{Intellectual Property}
	\acro{HT}{Hardware Trojan}
	\acro{ML}{Machine Learning}
	\acro{RL}{Reinforcement Learning}
	\acro{RTL}{Register Transfer Level}
	\acro{IC}{Integrated Circuit}	
	\acro{DUV}{Design Under Verification}
	\acro{DUT}{Design Under Test}
	\acro{EDA}{Electronic Design Automation}
	\acro{HDL}{Hardware Description Language}
	\acro{HSB}{Hardware Simulation Binary}
	\acro{AFL}{American Fuzzy Lop}
	\acro{ISA}{Instruction Set Architecture}
	\acro{CSR}{Control and Status Register}
	\acro{MDA}{Model Driven Architecture}
	\acro{UML}{Unified Modeling Language}
	\acro{NRE}{Non-Recurring Engineering}
	\acro{API}{Application Programming Interface}
	\acro{GUI}{Graphical User Interface}
	\acro{XML}{Extensible Markup Language}
	\acro{ToM}{Template of MetaFuzz}
	\acro{CSV}{Comma Separated Values}
	\acro{AES}{Advanced Encryption Standard}
	\acro{KMAC}{Keccak Message Authentication Code}
	\acro{HMAC}{Hash-based Message Authentication Code}
	\acro{RV-Timer}{RISC-V Timer}
	\acro{IPs}{Intellectual Properties}
	\acro{ToTB}{Template of Testbench}
	\acro{CWE}{Common Weakness Enumeration}
	\acro{PPO}{Proximal Policy Optimization}
	\acro{JTAG}{Joint Test Action Group}
	\acro{TAP}{Technology Access Port}
	\acro{CORDIC}{ COordinate Rotation Digital Computer}
	\acro{FIR}{ Finite Impulse Response}
	\acro{FIFO}{First-In, First-Out}
	\acro{RISC}{Reduced Instruction Set Computer}
	\acro{A2C}{Advantage Actor Critic}
	\acro{DQN}{Deep Q Network}
	\acro{ALU}{Arithmetic Logic Unit}
	\acro{DPI}{Direct Programming Interface}
	\acro{FSM}{Finite State Machine}
	\acro{SL}{Supervised Learning}
	\acro{ANN}{Artificial Neural Network}
	\acro{DNN}{Deep Neural Network}
	\acro{DT}{Decision Tree}
\end{acronym}


\title{Efficient Stimuli Generation using Reinforcement Learning in Design Verification \\

\thanks{This work has been developed in the project VE-VIDES (project label 16ME0243K) which is partly funded within the Research Programme ICT 2020 by the German Federal Ministry of Education and Research (BMBF)}
}

\ifthenelse{\boolean{blindreview}}{}{
	\author{\IEEEauthorblockN{
			Deepak Narayan Gadde\IEEEauthorrefmark{1},
			Thomas Nalapat\IEEEauthorrefmark{1},
			Aman Kumar\IEEEauthorrefmark{1}, \\
			Djones Lettnin\IEEEauthorrefmark{2},
			Wolfgang Kunz\IEEEauthorrefmark{3},
			Sebastian Simon\IEEEauthorrefmark{1}}
		\IEEEauthorblockA{
			\IEEEauthorrefmark{1}Infineon Technologies Dresden GmbH \& Co. KG, Germany \\
			\IEEEauthorrefmark{2}Infineon Technologies AG, Germany \\
			\IEEEauthorrefmark{3}Rheinland-Pf{\"a}lzische Technische Universit{\"a}t Kaiserslautern-Landau, Germany}
	}   
}


\maketitle
\pagenumbering{gobble}
\thispagestyle{firstpage}

\begin{abstract}
The increasing design complexity of \acp{SoC} has led to significant verification challenges, particularly in meeting coverage targets within a timely manner. At present, coverage closure is heavily dependent on constrained random and coverage driven verification methodologies where the randomized stimuli are bounded to verify certain scenarios and to reach coverage goals. This process is said to be exhaustive and to consume a lot of project time. In this paper, a novel methodology is proposed to generate efficient stimuli with the help of \ac{RL} to reach the maximum code coverage of the \ac{DUV}. Additionally, an automated framework is created using metamodeling to generate a SystemVerilog testbench and an \ac{RL} environment for any given design. The proposed approach is applied to various designs and the produced results proves that the \ac{RL} agent provides effective stimuli to achieve code coverage faster in comparison with baseline random simulations. Furthermore, various \ac{RL} agents and reward schemes are analyzed in our work.

\end{abstract}	


\section{Introduction} \label{sec:introduction}
\input{sections/introduction}

\section{Background Knowledge} \label{sec:background}
\input{sections/background}

\section{Related Work} \label{sec:related_work}
\input{sections/related_work}

\section{Proposed Approach} \label{sec:methodology}
\input{sections/methodology}

\section{Results} \label{sec:results}

\input{sections/results}


\section{Conclusion} \label{sec:conclusion}
\input{sections/conclusion}

\vspace{0.25cm}

\printbibliography

\end{document}

%% file: sections/introduction.tex
Due to the latest advancements in semiconductor technology, it is feasible to include various functionalities and features in a single \ac{SoC}. In the \ac{SoC} development, design verification continues to be one of the most expensive and time-consuming stages. A recent study done by the Wilson Research Group states that verification consumes around \SI{60}{\percent} of the overall project time \cite{VerStudy}. Hence, it is evident that the rising complexity of hardware designs necessitates the development of new approaches and methodologies which can provide verification engineers with the ability to fulfill their objectives faster and with minimal resources.

The simulation-based design verification is a well-established and powerful technique which utilizes \ac{CRV} and coverage-driven methodologies \cite{mehta2018asic}. Although it enhances the verification process, reaching coverage targets remains a greater challenge and is required for verification closure. This is primarily due to the demand for manual involvement of verification engineers adjusting constraints inside the testbench based on previous random simulations to drive the stimuli to reach the ultimate coverage objective.

Numerous initiatives have been made to optimize design verification through the use of \ac{ML} techniques. These are reviewed in \cite{siemens}, \cite{electronics10212688}, and \cite{mi13111887}. Most of these studies focused on functional coverage improvement, simulation speedup, and reducing the test count. This paper presents a novel approach to addressing the coverage closure problem in constrained random simulations by formulating it as an \ac{RL} task. Furthermore, it addresses how \ac{RL} could produce a better stimulus in comparison with a typical random stimulus during the simulation to achieve target code coverage of a given \ac{DUV}. This study aims to promote the adoption of automation methodologies that utilize metamodeling and \ac{RL} by showcasing the performance and comparing various \ac{RL} models and reward schemes employed in the approach.

The contributions of this work are as follows:
\begin{itemize}
	\item An novel method to reach code coverage goals faster with the help of \ac{RL} stimuli (Sec. \ref{sec:methodology})
	\item A metamodeling framework to build a generic SystemVerilog testbench for the given \ac{DUV} (Sec. \ref{subsec:metamodel})
	\item A configurable \ac{RL} environment to utilize various \ac{RL} policies and to create \ac{RL} actions for the given \ac{DUV} (Sec. \ref{subsec:rl_env})
	\item  The integration of the \ac{RL} environment into the simulation environment (Fig. \ref{framework_RL})
\end{itemize}

%% file: sections/background.tex
This section introduces the relevant techniques used in our work.

\subsection{Reinforcement Learning}
\begin{figure}[htb!]
	\centering
	\includegraphics[width=0.7\linewidth]{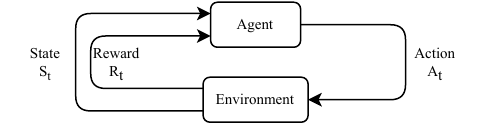}
	\caption{RL algorithm}
	\label{rl_process}
\end{figure}

\ac{RL} is a special type of \ac{ML} algorithm and is shown in Fig. \ref{rl_process}. Here, an agent interacts with an environment by performing certain actions $A_{t}$. For each action, the state $S_{t}$ of the agent gets altered and the agent gets a reward or penalty $R_{t}$ for action $A_{t}$ as feedback \cite{sutton}. The agent keeps performing these three tasks, which help it learn about and explore the environment around it. The agent develops insights on which actions result in rewards and which ones lead to negative feedback or penalties. It has several core components which are defined in the following Table \ref{rl_concept_table} along with their representation in our given coverage closure problem.

\begin{table}[htb!]
	\setcellgapes{0.5pt}
	\scriptsize	
	\makegapedcells
	\caption{\ac{RL} core components and their analogy in our work}
	\label{rl_concept_table}
	\centering
	\begin{tabular}{|l|l|l|}
		\hline
		\textbf{\begin{tabular}[c]{@{}l@{}}RL core \\ components\end{tabular}} & \textbf{Definition}                                                                                                     & \textbf{Analogy in our work}                                                                                       \\ \hline
		\textbf{Agent}                                                         & \begin{tabular}[c]{@{}l@{}}The learner or \\ decision maker\end{tabular}                                                & \begin{tabular}[c]{@{}l@{}}An intelligent unit\\ which generates a stimulus\end{tabular}                             \\ \hline
		\textbf{Environment}                                                   & \begin{tabular}[c]{@{}l@{}}The world with which \\ the agent interacts\end{tabular}                                     & \begin{tabular}[c]{@{}l@{}}A simulator which performs \\ RTL simulations\end{tabular}                              \\ \hline
		\textbf{Action}                                                        & \begin{tabular}[c]{@{}l@{}}All possible choices \\ the agent can take\end{tabular}                                      & New stimuli to drive DUV                                                                                           \\ \hline
		\textbf{Reward}                                                        & \begin{tabular}[c]{@{}l@{}}Immediate return from the \\ environment based on \\ certain actions\end{tabular}             & \begin{tabular}[c]{@{}l@{}}+1 or -1 or 0 \\ depends on performance \\ of action on environment\end{tabular}        \\ \hline
		\textbf{State}                                                         & \begin{tabular}[c]{@{}l@{}}The current situation returned \\ by the environment\end{tabular}                            & Current coverage score                                                                                             \\ \hline
		\textbf{Target}                                                        & The goal of the agent                                                                                                   & Reach 100\% coverage                                                                                               \\ \hline
		\textbf{Termination}                                                   & The state to end the RL process                                                                                         & \begin{tabular}[c]{@{}l@{}}When 100\% coverage is \\ reached or after certain \\ number of RL actions\end{tabular} \\ \hline
		\textbf{Policy}                                                        & \begin{tabular}[c]{@{}l@{}}The strategy that the agent \\ employs to determine its \\ actions at any state\end{tabular} & \begin{tabular}[c]{@{}l@{}}This work explores various \\ policies e.g., actor-critic\end{tabular}                  \\ \hline
	\end{tabular}

\end{table}

\subsection{Metamodeling}

\begin{figure}[htb!]
	\centering
	\includegraphics[width=0.7\linewidth]{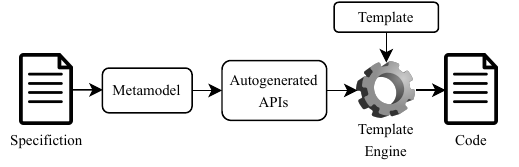}
	\caption{Metamodeling flow}
	\label{meta_flow}
\end{figure}

Code generation is crucial for boosting chip design productivity. However, Python-based code generators can lead to inconsistencies in data structures. Metamodeling \cite{schreiner2017new} helps ensure interoperability of data among various generators and promotes reusable code. It is essential for addressing inconsistencies across diverse systems with multiple viewpoints that require consistent automatic code generation from common sources. Fig. \ref{meta_flow} shows the basic flow in metamodeling.

In metamodeling, each model (design) has a corresponding \emph{metamodel} that defines its structure, constraints, and properties. These metamodels are usually represented as structured data sources like \emph{XML} files. A metamodeling environment facilitates interaction with these models by providing access, creation, and transformation capabilities based on the metamodel's description. The environment typically includes an automatically generated API, which allows users to interact with model instances and their properties. This API is utilized by a Mako-based \cite{mako} template and a template engine for generating the desired code. 

%% file: sections/related_work.tex
Over the past two decades, there has been a significant amount of research on \ac{ML} techniques to enhance the design verification process: \cite{siemens}, \cite{electronics10212688}, and \cite{mi13111887}. \ac{RL}, a subset of \ac{ML}, has also been investigated for solving various problems within the domain of verification. The first work to propose \ac{RL} in verification was \cite{hughes_optimizing_nodate}, where it was employed in combination with \ac{SL} to achieve higher functional coverage. The recent studies \cite{varambally2020optimising}, \cite{zheng_using_2022}, and \cite{ghany_speed_2021} utilized various \ac{SL} algorithms such as \ac{DNN}, \ac{ANN}, and \ac{DT} to significantly enhance simulation speed in comparison to random simulation approaches. Meanwhile, \ac{RL} is applied in the works \cite{bhuvaneswary_hybrid_2023}, \cite{10174129}, and \cite{ohana23} for the same purpose. Additionally, \cite{xu_reinforcement_2020}, \cite{choi_application_2021}, \cite{shibu_verlpy_2021}, and \cite{mi13111887} utilize \ac{RL} to reach target functional coverage more rapidly.

\begin{table}[]
	\setcellgapes{0.5pt}
	\scriptsize		
	\makegapedcells
	\caption{Comparision of related works}
	\label{related_table}
	\centering
	\begin{tabular}{|l|l|l|l|l|l|}
		\hline
		\textbf{Work}                                     & \textbf{Year} & \textbf{\begin{tabular}[c]{@{}l@{}}ML \\ Approach\end{tabular}} & \textbf{\begin{tabular}[c]{@{}l@{}}ML \\ Model\end{tabular}}    & \textbf{\begin{tabular}[c]{@{}l@{}}Testbench \\ Language\end{tabular}} & \textbf{Application} \\ \hline
		\cite{hughes_optimizing_nodate} & 2019          & \begin{tabular}[c]{@{}l@{}}SL,\\ RL\end{tabular}                & \begin{tabular}[c]{@{}l@{}}DNN,\\ Q-Learning\end{tabular}       & SV                                                              & FCI, SS                     \\ \hline
		\cite{varambally2020optimising}   & 2020          & SL                                                              & ANN                                                             & \begin{tabular}[c]{@{}l@{}}Python \\ Cocotb\end{tabular}        & SS                     \\ \hline
		\cite{pfeifer_reinforcement_2020}    & 2020          & RL                                                              & Q-Learning                                                      & -                                                               & FCI                     \\ \hline
		\cite{xu_reinforcement_2020}    & 2020          & RL                                                              & \begin{tabular}[c]{@{}l@{}}Tree search,\\ NN\end{tabular}       & -                                                               & FCI                     \\ \hline
		\cite{choi_application_2021}    & 2021          & RL                                                              & Custom                                                          & -                                                               & FCI, SS                  \\ \hline
		\cite{shibu_verlpy_2021}        & 2021          & RL                                                              & Soft Actor-Critic                                                             & \begin{tabular}[c]{@{}l@{}}Python \\ Cocotb\end{tabular}        & F                     \\ \hline
		\cite{mi13111887}  & 2022          & RL                                                              & SGTD                                                          & -                                                               & RTS                     \\ \hline
		\cite{zheng_using_2022}         & 2022          & SL                                                              & NN                                                              & -                                                           & FCI, SS                     \\ \hline
		\cite{ghany_speed_2021}         & 2022          & SL                                                              & \begin{tabular}[c]{@{}l@{}}ANN,\\ DNN,\\ DT\end{tabular} & SV-UVM                                                          & SS                     \\ \hline
		\cite{bhuvaneswary_hybrid_2023} & 2023          & RL                                                              & DNN                                                             & SV-UVM                                                          & SS                     \\ \hline
		\cite{liang2023late}            & 2023          & UL                                                              & 
		Isolation forest                                                & -                                                         & SS,TS
		\\ \hline
		\cite{10174129}            & 2023          & RL                                                              & 
		Actor-Critic                                                & SV-UVM                                                          & FCI
		\\ \hline
		\cite{ohana23}            & 2023          & RL                                                              & 
		DQN                                                & SV                                                          & FCI
		\\ \hline
		Our work            	  & 2024          & RL                                                              & 
		PPO, A2C, DQN                                                & SV                                                          & CCI
		\\ \hline
	\end{tabular}
	\begin{center}
		\vspace{1ex}
		\justifying
		\scriptsize  Notes: Unsupervised Learning (UL), Simulation Speedup (SS), Functional Coverage Improvement/Closure (FCI), Code Coverage Improvement/Closure (CCI), Reaching Target State (RTS), Test Selection (TS), SystemVerilog (SV), Universal Verification Methodology (UVM)
	\end{center}
\end{table}

Our work is compared with other relevant research based on various aspects to illustrate its significance as depicted in Table \ref{related_table}. Most of the works which employed an \ac{RL} approach require manual effort to configure the \ac{RL} environment and to write a design-specific testbench. These works utilized specific \ac{RL} algorithms in their methods and did not examine how other \ac{RL} models would perform for the same application. The approach presented in this paper is design-agnostic and configurable in terms of learning policy, reward scheme, and target coverage type. Additionally, we produced the results by employing three \ac{RL} algorithms in our approach, namely \ac{A2C}, \ac{PPO}, and \ac{DQN} from the Stable-Baselines3 library \cite{stable-baselines3}, to show their performance to reach maximum code coverage of the \ac{DUV}. These algorithms were chosen based on their support for continuous and discrete actions. \ac{A2C} and \ac{PPO} support both types of action space, whereas \ac{DQN} supports only the discrete type.

%% file: sections/methodology.tex
\begin{figure}[h!]
	\centering
	\includegraphics[width=0.8\linewidth]{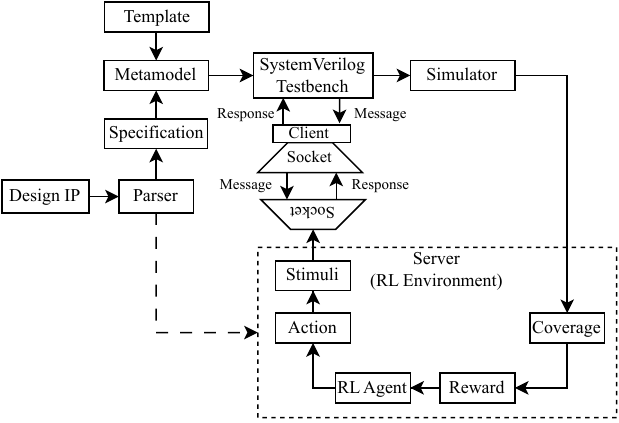}
	\caption{\ac{RL}-guided design verification}
	\label{framework_RL}
\end{figure}

Fig. \ref{framework_RL} depicts the proposed framework, which begins by parsing the design \ac{IP} to extract all information about the primary ports. This data is used by the metamodel to generate a SystemVerilog testbench that interfaces with an \ac{RL} environment implemented in Python. The simulation begins with the \ac{RL}  environment sending initial zero values for all ports to the testbench. Subsequently, the \ac{DUV} is simulated using the values from the \ac{RL} environment. Coverage for this timestep is collected using the simulator, which is then relayed back to the \ac{RL} environment to calculate the reward. Based on the reward and coverage value, the \ac{RL} agent selects a new action i.e., a new stimulus which will in turn be used to stimulate the \ac{DUV} for another step. This process is repeated until the coverage reaches \SI{100}{\percent} or until after a predetermined number of steps.

The following subsections provide a detailed explanation of these processes:

\subsection{Parsing and specification formation}
The \ac{RTL} code of the given design \ac{IP}, usally in VHDL, Verilog, or SystemVerilog, is parsed to extract the primary port parameters such as name, type, size, and direction of the ports. This information is used to create a standardized specification in a \ac{XML} format. 

\subsection{Testbench creation} \label{subsec:metamodel}
The metamodel utilizes this specification and a predefined template to create a testbench in SystemVerilog. The \ac{DUV} must be driven by the stimuli received from the RL agent. Direct communication with the SystemVerilog testbench is not feasible as our \ac{RL} environment is implemented in Python.

This communication can be achieved by implementing a client-server application \cite{Bob_2019}. The \ac{DPI} \cite{mehta2018asic} can be utilized to call functions written in C from the SystemVerilog testbench. Therefore, a client responsible for creating a client socket, managing handshakes, and sending the response from the server, which in this context is the RL environment, is implemented in C.

Based on a predefined template, the metamodel creates a testbench that requests stimuli from the server and drives the inputs of the \ac{DUV} with the received data.

\subsection{Simulation and coverage collection}
The simulator is configured to dump coverage data after each clock cycle. The specific type of coverage (block, \ac{FSM}, toggle, or expression) is determined by the settings in the configuration file. The \ac{RL} environment reads the saved coverage value after each clock cycle.

\subsection{RL environment} \label{subsec:rl_env}
The \ac{RL} environment based on OpenAI's Gym \cite{brockman2016openai} is configured by providing the below details in the configuration file.
\begin{itemize}
	\item \emph{Top Module} - The design IP to be verified
	\item \emph{Coverage Type} - The type of coverage (block, \ac{FSM}, toggle or expression) based on which the reward is calcluated
	\item \emph{Learning Policy} - \ac{RL} algorithm (\ac{PPO}, \ac{A2C} or \ac{DQN}) that guides the learning process of the \ac{RL} agent
	\item \emph{Ports} - The ports of the \ac{IP} that contribute to the increase in coverage
	\item  \emph{Reward Scheme} - Optimistic or penalty based reward scheme
\end{itemize}

\subsubsection{Actions}
The action space is defined by the designated ports. The \ac{RL} environment reads the size of these ports from the parser's output and configures the action space to encompass all possible combinations of values that the ports can assume. For instance, for an \ac{ALU} with a 3-bit opcode, the action space would range from 0 to $2^3-1$, and action 6 would drive the \emph{opcode} port of the ALU with its binary value \emph{110}.

\subsubsection{Reward}
Following each clock cycle of the simulation, the \ac{RL} environment reads the coverage data dumped by the simulator. If the coverage at the current step exceeds the coverage from the previous step, the reward is assigned a value of 1. If the coverage decreases, the reward is determined according to the predefined reward scheme. Algorithm \ref{reward_scheme} illustrates the reward calculation process for both optimistic and penalty-based reward schemes.

\begin{algorithm}[htb!]
\footnotesize
\caption{Reward Scheme used in our framework}
\label{reward_scheme}
\begin{algorithmic}[1]
	\IF{current\_coverage > previous\_coverage}
	\STATE reward $\gets$ 1
	\ELSIF{reward\_scheme == penalty}
	\STATE reward $\gets$ -1
	\ELSIF{reward\_scheme == optimistic}
	\STATE reward $\gets$ 0
	\ENDIF
	
\end{algorithmic}
\end{algorithm}

\subsubsection{RL agent}
The \ac{RL} agent selects an action according to the specified policy. It then performs the chosen action by packaging the stimulus and port name into a message packet and sending this packet to the client (SystemVerilog testbench) upon receiving a request. Subsequently, the agent receives a reward corresponding to this action. This reward assesses the effectiveness of the action, and the agent updates its policy and chooses the next action based on this received reward. The chosen action is performed in the next step. This process continues until coverage reaches \SI{100}{\percent} or until after a predetermined number of steps.

\begin{table*}[ht!]
	\setcellgapes{1pt}
	\scriptsize
	\makegapedcells
	\caption{Results produced on various design IPs to reach maximum code coverage}
	\label{result_table}
	\centering
	\begin{tabular}{ccccccccc}
		\multirow{3}{*}{\textbf{Design IP}} & \multirow{3}{*}{\textbf{\begin{tabular}[c]{@{}c@{}}Max. Code \\ Coverage\\ (\%)\end{tabular}}} & \multirow{3}{*}{\textbf{\begin{tabular}[c]{@{}c@{}}\#Random\\ Stimuli\end{tabular}}} & \multicolumn{6}{c}{\textbf{\#RL Stimuli}}                                                                                                                                                                                                                                                                                                                 \\ \cline{4-9} 
		&                                                                                           &                                                                                      & \multirow{2}{*}{\textbf{\begin{tabular}[c]{@{}c@{}}PPO\\ (Optimistic)\end{tabular}}}  & \multirow{2}{*}{\textbf{\begin{tabular}[c]{@{}c@{}}PPO\\ (Penalty)\end{tabular}}} & \multirow{2}{*}{\textbf{\begin{tabular}[c]{@{}c@{}}A2C\\ (Optimistic)\end{tabular}}}  & \multirow{2}{*}{\textbf{\begin{tabular}[c]{@{}c@{}}A2C\\ (Penalty)\end{tabular}}} & \multirow{2}{*}{\textbf{\begin{tabular}[c]{@{}c@{}}DQN\\ (Optimistic)\end{tabular}}}  & \multirow{2}{*}{\textbf{\begin{tabular}[c]{@{}c@{}}DQN\\ (Penalty)\end{tabular}}} \\
		&                                                                                           &                                                                                      &                               &                                                                                   &                               &                                                                                   &                               &                                                                                   \\ \hline
		\textbf{JTAG TAP}
		\cite{ambuja}                   & 94.39                                                                                     & 1699                                                                                 & 1373                          & 545                                                                               & 627                           & 1225                                                                              & 493                           & 975                                                                               \\
		\textbf{ALU}                        & 90.91                                                                                     & 22                                                                                   & 12                            & 8                                                                                 & 16                            & 22                                                                                & 27                            & 20                                                                                \\
		\textbf{CORDIC} \cite{opencores}                     & 99.72                                                                                     & 124                                                                                  & 95                            & 55                                                                                & 105                           & 63                                                                                & 60                            & 96                                                                                \\
		\textbf{RISC-V} \cite{opencores}                     & 84.58                                                                                     & 330                                                                                  & 226                           & 291                                                                               & 506                           & 535                                                                               & 734                           & 271                                                                               \\
		\textbf{FIR} \cite{ambuja}                          & 100                                                                                       & 11                                                                                   & 6                             & 10                                                                                & 22                            & 11                                                                                & 10                            & 6                                                                                 \\
		\textbf{FIFO}
		\cite{ambuja}                       & 100                                                                                       & 23                                                                                   & 22                            & 23                                                                                & 22                            & 22                                                                                & 22                            & 31                                                                               
	\end{tabular}
\end{table*}

%% file: sections/results.tex
In this section, we present the results obtained by applying stimuli generated by the \ac{RL} agent to various designs. Subsequently, we compare the number of steps required to achieve maximum code coverage using stimuli generated randomly and those generated by \ac{RL} agents based on different policies. 

Table \ref{result_table} presents a comparison of the number of stimuli generated randomly versus those produced by \ac{RL} agents with various policies and reward schemes to reach maximum code coverage across all designs. It is evident that \ac{RL} agents require fewer stimuli to attain the maximum possible code coverage when compared with random stimuli. For the majority of the designs, \ac{RL} agents based on \ac{PPO} yield superior results with the penalty scheme. However, the \ac{PPO} agent with the optimistic scheme reached the same final coverage of RISC-V and FIR design \acp{IP} with fewer simulations.

\begin{figure}[htb!]
	\centering
	\includegraphics[width=\linewidth]{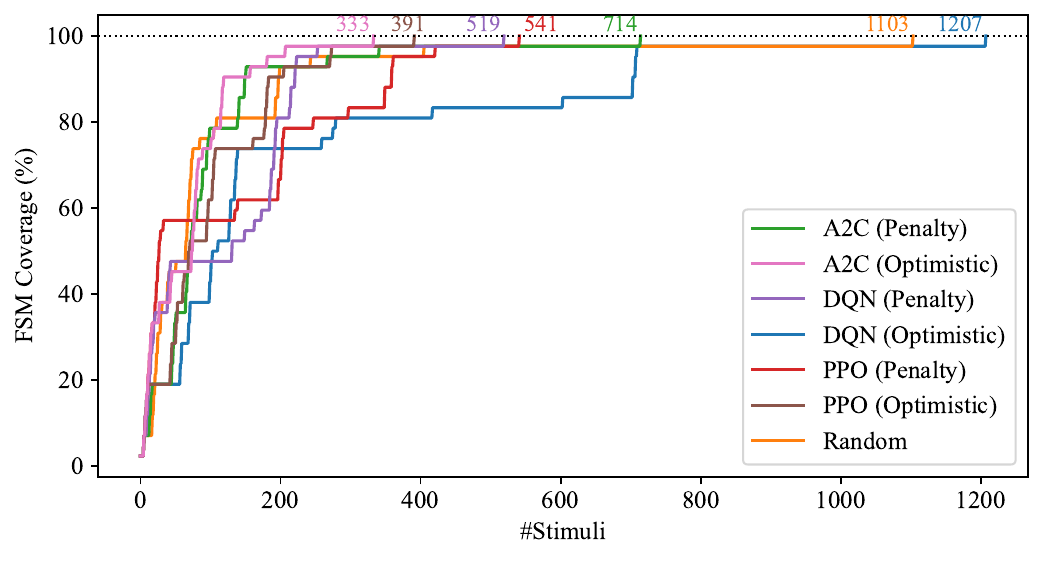}
	\caption{Progression of the FSM Coverage with respect to stimuli generated by different RL agents and random simulation}
	\label{graph_JTAG}
\end{figure}

An additional experiment was done on JTAG TAP \ac{IP} to reach maximum \ac{FSM} coverage to see how \ac{RL} agents perform if we try to target a specific coverage type. Fig. \ref{graph_JTAG} illustrates the change in \ac{FSM} coverage for \ac{RL} stimuli generated by different agents as well as random stimuli. Our results show that \ac{RL} agents were able to achieve the maximum code coverage with a lower number of stimuli. The \ac{A2C} agent with the optimistic reward scheme outperformed other agents by reaching \SI{100}{\percent} \ac{FSM} coverage around 3 times faster than random stimulation. 

%% file: sections/conclusion.tex
In this work, we introduce a design-agnostic framework leveraging \ac{RL} for design verification purposes. By automating the setup of \ac{RL} environments and generating customized testbenches according to the design being verified, our framework streamlines the verification process. Results obtained from verifying six distinct designs confirm that \ac{RL}-guided verification requires a reduced number of stimuli relative to conventional random simulations to attain the threshold coverage. Moreover, our findings indicate that the \ac{PPO}-based \ac{RL} agent often exhibits superior performance compared to \ac{DQN} and \ac{A2C} agents. 